\documentclass[conference]{IEEEtran}
\IEEEoverridecommandlockouts
\usepackage{cite}
\usepackage{amsmath,amssymb,amsfonts}
\usepackage{algorithmic}
\usepackage{graphicx}
\usepackage{textcomp}
\usepackage{xcolor}
\usepackage{subcaption}
\usepackage{hyperref}

\usepackage{multirow}
\usepackage{acronym}
\usepackage[ruled]{algorithm2e}
\makeatletter
\newcommand{\algrule}[1][.2pt]{\par\vskip.5\baselineskip\hrule height #1\par\vskip.5\baselineskip}
\makeatother

\def\BibTeX{{\rm B\kern-.05em{\sc i\kern-.025em b}\kern-.08em
    T\kern-.1667em\lower.7ex\hbox{E}\kern-.125emX}}

\DeclareMathOperator{\abs}{abs}
\DeclareMathOperator{\sqr}{sqr}
\DeclareMathOperator{\fold}{fold}
\DeclareMathOperator{\dist}{dist}
\DeclareMathOperator{\mean}{mean}
\DeclareMathOperator{\std}{std}
\DeclareMathOperator{\Lone}{L1}
\DeclareMathOperator{\Ltwo}{L2}

\begin{document}

\def\input{\mathbf{x}}

\newacro{AAKELM}[AAKELM]{Autoassociative Kernelized Extreme Learning Machine} 
\newacro{CNN}[CNN]{Convolutional Neural Network}
\newacro{GESSVDD}[GESSVDD]{Graph-embedded Subspace Support Vector Data Description}
\newacro{GRM}[GRM]{Generalized Reference Mapping}
\newacro{ELM}[ELM]{Extreme Learning Machine}
\newacro{ESVDD}[ESVDD]{Ellipsoidal Support Vector Data Description}
\newacro{KDA}[KDA]{Kernel Discriminant Analysis}
\newacro{KPCA}[KPCA]{Kernel Principal Component Analysis}
\newacro{LDA}[LDA]{Linear Discriminant Analysis}
\newacro{OCC}[OCC]{One-class Classification}
\newacro{OCELM}[OCELM]{One-class Extreme Learning Machine}
\newacro{OCKELM}[OCKELM]{One-class Kernelized Extreme Learning Machine}
\newacro{OCSVM}[OCSVM]{One-class Support Vector Machine}
\newacro{NPT}[NPT]{Non-linear Projection Trick}
\newacro{PCA}[PCA]{Principal Component Analysis}
\newacro{RBF}[RBF]{Radial Basis Function}
\newacro{REF}[REF]{Repeated Element-wise Folding}
\newacro{SVDD}[SVDD]{Support Vector Data Description}
\newacro{SSVDD}[SSVDD]{Subspace Support Vector Data Description}
\newacro{GSVDD}[GSVDD]{Generalized Support Vector Data Description}
\newacro{SVM}[SVM]{Support Vector Machine}

\title{Linear-time One-Class Classification with\\
Repeated Element-wise Folding\\
}

\author{\IEEEauthorblockN{Jenni Raitoharju}
\IEEEauthorblockA{\textit{Faculty of Information Technology}, 
\textit{University of Jyväskylä},
Jyväskylä, Finland,
0000-0003-4631-9298}
}

\maketitle

\begin{abstract}
This paper proposes an easy-to-use method for one-class classification: \acf{REF}. The algorithm consists of repeatedly standardizing and applying an element-wise folding operation on the one-class training data. Equivalent mappings are performed on unknown test items and the classification prediction is based on the item's distance to the origin of the final distribution. 
As all the included operations have linear time complexity, the proposed algorithm provides a linear-time alternative for the commonly used computationally much more demanding approaches. Furthermore, \ac{REF} can avoid the challenges of hyperparameter setting in one-class classification by providing robust default settings. The experiments show that the proposed method can produce similar classification performance or even outperform the more complex algorithms on various benchmark datasets. Matlab codes for \ac{REF} are publicly available at \href{https://github.com/JenniRaitoharju/REF}{https://github.com/JenniRaitoharju/REF}.
\end{abstract}


\section{Introduction}

\ac{OCC} is used in situations, where it is necessary to create a model for outlier detection, while no outlier samples are available for training the model, but only normal/target data. \ac{OCC} methods are traditionally divided into density-based (e.g., Parzen \cite{duin1976choice}), boundary-based (e.g., \ac{OCSVM} \cite{scholkopf1999support}, \ac{SVDD} \cite{tax2004support}), and reconstruction-based (e.g., \ac{AAKELM} \cite{gautam2017construction})  algorithms. Probably the most well-known approaches are the traditional support vector methods, \ac{OCSVM} and \ac{SVDD}, with several extensions (e.g., \cite{sohrab2021multimodal,wu2023adaptive}) and applications (e.g., \cite{sohrab2020boosting, al2020unsupervised}) appearing regularly. A major limitation of these methods is their cubic time complexity \cite{sohrab2021multimodal}. \ac{ELM}-based methods provide faster solutions, but they perform better as kernel versions \cite{leng2015one}, which leads to quadratic complexities. As a result, many common \ac{OCC} algorithms do not scale up well. In recent years, also deep learning has been used in \ac{OCC} \cite{ruff2018deep,goyal2020drocc}. With deep learning-based methods, the challenge is somewhat opposite as they require a lot of data for learning representative features. The methods are not suitable for small datasets where the traditional methods are still regularly used. 

Hyperparameter settings often significantly affect the performance of \ac{OCC} methods \cite{sohrab2021multimodal}.
As the assumption is that only target class data are available for training, selecting optimal hyperparameter values is a special challenge. One approach for selecting the hyperparameter values is consistency-based model selection \cite{tax2004consistency}, where the main idea is to increase the complexity of the classifier until it becomes inconsistent on the target class. However, this approach is based on preselecting the fraction of target class items that should be rejected by the model. This fraction of rejection is also an impactful hyperparameter and fixing it to a certain value will likely lead to sub-optimal results. Another commonly used approach (used e.g., in \cite{sohrab2021multimodal,sohrab2023graph}) is to assume that some amount of outlier training data is available for hyperparameter selection, while this data is not used for the actual model training. Both approaches can be very time-consuming especially on large datasets and in the case of multiple hyperparameters. Implementation of the hyperparameter optimization process also requires some skills and makes the approaches more cumbersome to use. Therefore, a hyperparameter-free method would provide a desirable solution. 

This paper complements the repertoire of \ac{OCC} methods with an easy-to-implement, simple-to-use, linear-time method, which performs well also without any hyperparameter tuning.

\section{One-class Classification with Repeated Element-wise Folding}

A very simple approach for \ac{OCC} would be to standardize the data and then evaluate test samples based on their distance to the center of the data. The challenge is deciding a suitable threshold for classification, but if some outliers are available for hyperparameter selection, the approach is surprisingly competitive compared to more complex \ac{OCC} approaches (as shown by the \emph{base} approach in the experiments).

To avoid the challenge of selecting the classification threshold, we consider the following: Approximately 68\% of normally distributed data fall within one standard deviation. Taking the absolute value of the distribution centered to origin results in a half normal distribution, a special case of folded normal distribution \cite{tsagris2014folded}. As the half normal distribution has a narrower spread around its mean, a larger part of the data falls within one standard deviation. It can be easily experimentally shown that continuing the process by \emph{alternatingly centering the data to the origin and taking the absolute value}, the fraction of data falling within one standard deviation gradually grows close to 100\%. At the same time, data that originally lied outside the distribution (further away from the mean than any of the data points) will remain outside one standard deviation. 

%This motivates a simple essentially parameter-free one-class classifier that repeatedly standardizes and folds the data by taking the absolute value. The standardization is both the first and the last step in the series. For testing, the mean and standard deviation values used in the standardization steps are saved and then applied to the unknown test item. All samples within one standard deviation are classified as target class, others as outliers.   

\begin{figure}[t!]
\center{\includegraphics[width=0.7\columnwidth]{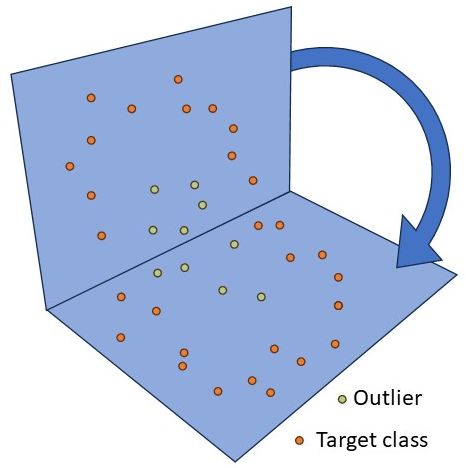}}
\caption{Simplified illustration of element-wise folding. Data distribution may have cavities (potential outlier areas). In the example, folding along both dimensions and standardizing again could help moving the outliers outside the target class distribution. }
\vspace{-5px}
\label{fig:fold}
\end{figure}

In real-life \ac{OCC} tasks, the target data is not necessarily normally distributed, but the same phenomenon can be observed also for different data distributions. The question is what happens for outliers that initially lie closer to the mean than some of the target class samples. If the outlier data are simply overlapping with the target distribution,  any \ac{OCC} approach would fail. However, the proposed approach  has potential to remove some cavities (potential outlier regions) appearing inside the target distribution. Applying a folding operation may help to move the cavity from the middle of the distribution to the side illustrated in Fig.~\ref{fig:fold}. When the data are then standardized again, the target distribution becomes more uniform. Similarly, if the target distribution consists of several distinct clusters, the folding operation may map them together. Once outliers fall outside one standard deviation of the target distribution, they will stay there in the later iterations. Naturally, some outlier regions may be asymmetrically located so that folding makes them more entangled with the target data. Nevertheless, such regions would have been misclassified also without the folding operations, and such scenarios would be challenging for any \ac{OCC} approach. 

\begin{algorithm}[t!]
\SetAlgoLined
\caption{Repeated Element-wise Folding One-Class Classifier }
\SetAlgoLined
\vspace{6pt}
\textbf{Training}\\
\algrule
\SetKwInOut{Input}{Input}
\SetKwInOut{Output}{Output}
\Input{$\mathbf{X} = \{\mathbf{x_1}, \dots, \mathbf{x_N}\} \in \mathbb{R}^D$, \%  \emph{OC train data}\\
$J$, \% \emph{Number of iterations (default 101)}\\
$\fold()$, \% \emph{Folding operation (default $\abs$)}\\}

\vspace{2mm}
\Output{$\mathbf{M}= \{\mathbf{\mu_1}, \dots, \mathbf{\mu_J}\} \in \mathbb{R}^D$, \% \emph{Mean vectors} \\
$\mathbf{S}= \{\mathbf{\sigma_1}, \dots, \mathbf{\sigma_J}\} \in \mathbb{R}^D$, \% \emph{Std vectors}}
 \vspace{3mm}

\% \emph{Standardize data}\\
$\mu_{1} = \mean( \mathbf{x_1}, \dots, \mathbf{x_N})$, $\sigma_{1} = \std( \mathbf{x_1}, \dots, \mathbf{x_N})$\\
\For{$n=1:N$}{
$\mathbf{x_n} = \mathbf{x_n}-\mathbf{\mu_{1}}$, $\mathbf{x_n} = \mathbf{x_n}./\mathbf{\sigma_{1}}$\\
}  
 
\For{$i=2:J$}{
  \% \emph{Element-wise folding}\\
  \For{$n=1:N$}{
    $\mathbf{x_n} = \fold(\mathbf{x_n})$\\} 
    
  \% \emph{Standardize data}\\
  $\mu_{i} = \mean( \mathbf{x_1}, \dots, \mathbf{x_N}),\sigma_{i} = \std( \mathbf{x_1}, \dots, \mathbf{x_N})$\\
  \For{$n=1:N$}{
    $\mathbf{x_n} = \mathbf{x_n}-\mathbf{\mu_{i}}$,
    $\mathbf{x_n} = \mathbf{x_n}./\mathbf{\sigma_{i}}$\\
    }  

}
\algrule
\textbf{Testing}\\
\algrule
\vspace{5pt}
\SetKwInOut{Input}{Input}
\SetKwInOut{Output}{Output}
\Input{$\mathbf{y}$, \%  \emph{Test sample}\\
$\mathbf{M}= \{\mathbf{\mu_1}, \dots, \mathbf{\mu_J}\} \in \mathbb{R}^D$, \% \emph{Mean vectors} \\
$\mathbf{S}= \{\mathbf{\sigma_1}, \dots, \mathbf{\sigma_J}\} \in \mathbb{R}^D$, \% \emph{Std vectors}\\
$T$, \% \emph{Classification threshold (default 1)} \\
$\dist()$, \% \emph{Distance metric (default $\Lone/D$)}\\}
\vspace{2mm}
\Output{$l = target/outlier$ \% \emph{class label for $\mathbf{y}$}}
\vspace{3mm}

\% \emph{Standardize $\mathbf{y}$ wrt. training data}\\
$\mathbf{y} = \mathbf{y}-\mathbf{\mu_{1}}$, $\mathbf{y} = \mathbf{y}./\mathbf{\sigma_{1}}$\\

\For{$i=2:J$}{

  \% \emph{Element-wise folding}\\
    $\mathbf{y} = \fold(\mathbf{y})$\\
    
  \% \emph{Standardize $\mathbf{y}$ wrt. training data}\\
    $\mathbf{y} = \mathbf{y}-\mathbf{\mu_{i}}$,
    $\mathbf{y} = \mathbf{y}./\mathbf{\sigma_{i}}$\\
}
\eIf{$\dist(\mathbf{y,0}) \leq T$}{$l=target$}{$l=outlier$}
\algrule
The Matlab implementation for \ac{REF} is available at \href{https://github.com/JenniRaitoharju/REF}{https://github.com/JenniRaitoharju/REF}.
\vspace{6pt}
\end{algorithm}

The resulting new \ac{OCC} approach is coined \acf{REF} and it is described in detail in Algorithm~1. Besides the default absolute value operation, other element-wise folding operations can be used in the algorithm as the fold() operation. Here, \emph{element-wise folding} refers to operations that map values from different parts of the input space together leaving some parts of the space empty. While operations such as $\cos$ or $\sin$ can be also considered as folding operations that map all the input values between -1 and 1, the main focus is on operations that create a single folding similar to the absolute value operation illustrated in Fig.~\ref{fig:fold}. In standardized data, the elements are within one standard deviation if their values are between -1 and 1. To approximately evaluate if this is the case, the default \ac{REF} computes the L1 distance of a test sample to the origin and divides it by the number of dimensions. Besides the default distance metric L1/$D$ and classification threshold 1, other distance metrics, dist(), and thresholds, $T$, can be considered. For example, if outliers are known to be overlapping with the target data distribution, a smaller threshold may lead to a better result. The default number of iterations, $J$, is set to $101$. It is sufficient to bring around 99.5\% of normally distributed target data within one standard deviation.

\section{Experiments}

\subsection{Datasets and Experimental Setup}

Experiments were conducted on six small datasets from UCI Machine Learning Repository (\href{http://archive.ics.uci.edu/ml}{http://archive.ics.uci.edu/ml}). The dataset properties are summarized in Table~\ref{tab:datasets}. A separate one-class classification task was created for each class in each dataset using samples from this class as the target data and the remaining samples as outliers. 70\% of the data were randomly selected for training and the remaining data for testing. This splitting was repeated five times and the mean and standard deviation of the performance over these splittings are reported. 
%For MNIST, the original train-test splitting was used. In each task, only target class training data was used for training, but the full test set was used for testing. 
All data were standardized wrt. training data for all methods.

All experiments were performed using Matlab R2017b.
I conducted experiment both using the hyperparameter optimization process including outlier training data (as in \cite{sohrab2021multimodal,sohrab2023graph}) and by using fixed default hyperparameters. When optimizing hyperparameters, random 5-fold cross-validation over the training set items was used to set the hyperparameter values for each task and train-test splitting separately.

Gmean was used as the evaluation metric. It is defined as the geometric mean over True Positive Rate (TPR) and True Negative Rate (TNR), i.e., $\text{Gmean}=\sqrt {\text{TPR} \times \text{TNR}}.$

\subsection{Model Verification}

Different implementation choices for \ac{REF} were first evaluated to validate the default settings. The following element-wise mappings were compared: absolute value ($\abs$), square ($\sqr$), cosine ($\cos$), sine ($\sin$), hyperbolic tangent ($\tanh$), and a combination of $\cos$ and $\abs$ ($\cos$-$\abs$), where cosine was used for $-1 \leq \mathbf{x}_i \leq 1$ and $\abs$ elsewhere. Out of these options, $\abs$, $\sqr$, and $\cos$-$\abs$ are folding operations that create a single folding as in Fig.~\ref{fig:fold}, $\cos$ and $\sin$ create multiple foldings, whereas $\tanh$ is not considered a folding operation here. The operations $\cos$-$\abs$ and $\cos$ are equivalent for the central part of the data ($-1 \leq \mathbf{x}_i \leq 1$), but differ for $|\mathbf{x}_i| > 1$. The compared distance metrics were $\Lone/D$ and $\Ltwo/D$. The number of iterations was set to $J=101$ in all the experiments. The classification threshold was either selected from $T =\{0.3, 0.4, 0.5, 0.6, 0.7, 0.8, 0.9, 1.0, 1.1\}$  or fixed to the default value $T=1$. The simple baseline approach (base), where the standardized items were directly classified based on their distance to the origin without any element-wise folding, was also evaluated. For this approach, the classification threshold $T$ was selected as in the other variants. Clearly, any \ac{REF} variant that does not outperform the base approach is useless in terms of improving the classification performance.  

The results are given in Table~\ref{tab:REFvariants}. It can be observed that the operations creating a single folding ($\abs$, $\sqr$, and $\cos$-$\abs$) indeed outperform the other operations, the default operation $\abs$ being the best. Compared to the base approach, the other operations make the results worse. As $\cos$ is similar to $\cos$-$\abs$ near the origin, it is not as bad as $\sin$ or $\tanh$. The results obtained for the well-performing operations are quite similar for the two distance metrics, but the default metric  $\Lone/D$ gives clearly better results for $\cos$. When the default \ac{REF} approach is compared to the base approach, it can be seen that the base approach is more dependent on hyperparameter optimization and the performance gap between the methods becomes much bigger when fixed default hyperparameters are used.

\begin{figure*}[bt]
    \centering
    \begin{subfigure}[t]{0.45\columnwidth}
        \centering
        \includegraphics[height=1.3in]{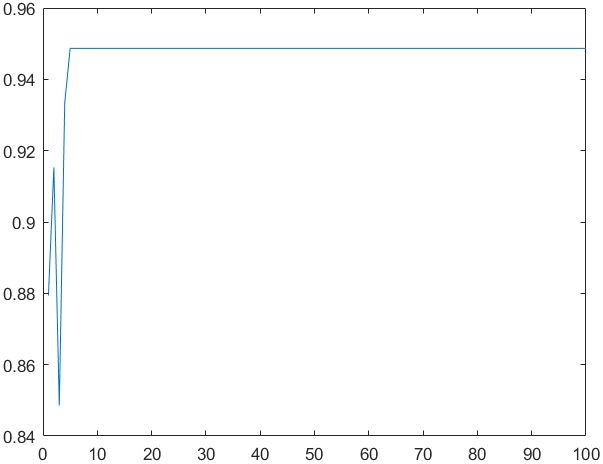}
        \caption{Iris, class 2}
        \label{sfig:iris}
    \end{subfigure}%
    \hfill
    \begin{subfigure}[t]{0.45\columnwidth}
        \centering
        \includegraphics[height=1.3in]{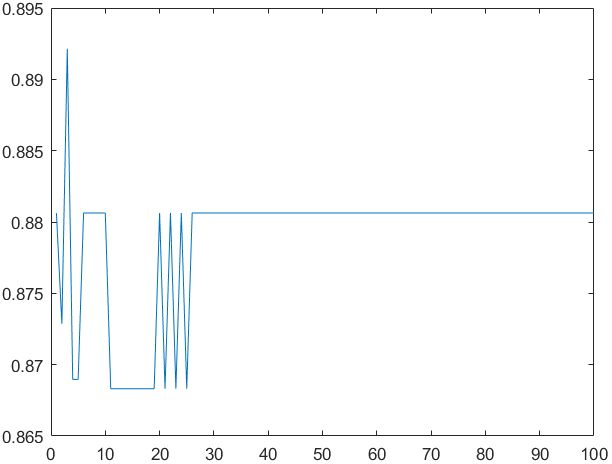}
        \caption{Seeds, class 1}
        \label{sfig:seeds}
    \end{subfigure}
    \hfill
    \begin{subfigure}[t]{0.45\columnwidth}
        \centering
        \includegraphics[height=1.3in]{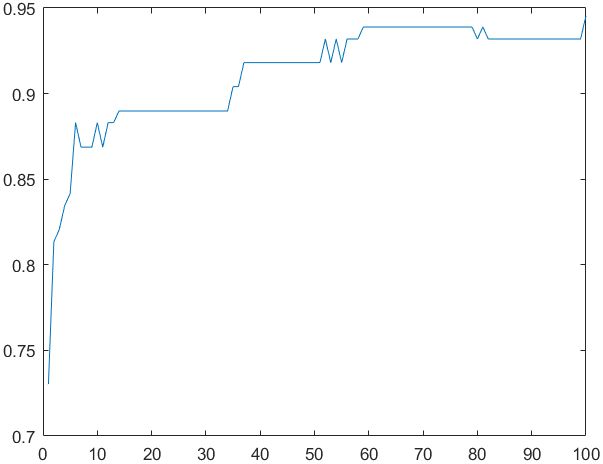}
        \caption{Ionosphere, class 1}
        \label{sfig:ionosphere}
    \end{subfigure}
    \hfill
    \begin{subfigure}[t]{0.45\columnwidth}
        \centering
        \includegraphics[height=1.3in]{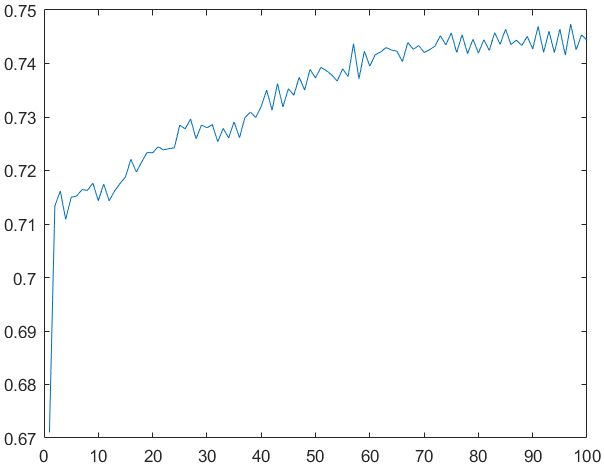}
        \caption{Aver. of cases w. $T=1$}
        \label{sfig:average}
    \end{subfigure}
    \caption{Different learning curves for \ac{REF} runs. Gmean vs. iteration}
    \label{fig:curves}
\end{figure*}

Fig.~\ref{fig:curves} shows performance curves over iterations for different \ac{REF} experiments. Figs.~\ref{sfig:iris}-\ref{sfig:ionosphere} show curves for an individual task and train-test splitting with optimized hyperparameters. It can be seen that sometimes the performance improvement is obtained early and no further changes occur as in Fig.~\ref{sfig:iris}, sometimes applying \ac{REF} does not help at all or even makes the results worse as in Fig.~\ref{sfig:seeds}, sometimes the performance improves gradually as in Fig.~\ref{sfig:ionosphere}. Fig.~\ref{sfig:average} is the average curve over all 14 tasks and 5 splittings (70 cases) with the default threshold $T=1$, i.e., it shows how the average result improves from the base approach (last column in Table~\ref{tab:REFvariants}) to the final \ac{REF} performance (second last column in Table~\ref{tab:REFvariants}) first rapidly and then gradually over the remaining iterations.

\begin{table}[tbp]
\caption{Datasets and subtasks used in the experiments \vspace{-5pt}}
\label{tab:datasets}
\begin{center}
\resizebox{0.95\linewidth}{!}{
\begin{tabular}{|c|cccccc|}
\hline
\bf{Dataset}& $C$ & $N_{tot}$ & $D$ & Task abr. & Target class & $N$ \\
\hline
\multirow{3}{*}{Iris} & \multirow{3}{*}{3} & \multirow{3}{*}{150} & \multirow{3}{*}{4} & Iris1 & Setosa & 35\\
&&&& Iris2 & Versicolor & 35\\
&&&& Iris3 & Virginica & 35\\
\hline
\multirow{3}{*}{Seeds} & \multirow{3}{*}{3} & \multirow{3}{*}{210} & \multirow{3}{*}{7} & Seed1 & Kama & 49\\
&&&& Seed2 & Rosa & 49\\
&&&& Seed3 & Canadian & 49\\
\hline
\multirow{2}{*}{Ionosphere} & \multirow{2}{*}{2} & \multirow{2}{*}{351} & \multirow{2}{*}{32} & Ion1 & Good & 157 \\
&&&& Ion2 & Bad & 88\\
\hline
\multirow{2}{*}{Sonar} & \multirow{2}{*}{2} & \multirow{2}{*}{208} & \multirow{2}{*}{60} & Son1 & Rock & 67 \\
&&&& Son2 & Mines & 77 \\
\hline
Qualitative  & \multirow{2}{*}{2} & \multirow{2}{*}{250} & \multirow{2}{*}{6} & Bank1 & No bankr. & 100\\
bankruptcy &&&& Bank2 & Bankr. & 74\\
\hline
Somerville  & \multirow{2}{*}{2} & \multirow{2}{*}{143} & \multirow{2}{*}{6} & Happ1 & Unhappy & 46 \\
happiness &&&& Happ2 & Happy & 53\\
\hline
\multicolumn{7}{l}{\vspace{-5pt}}\\
\multicolumn{7}{l}{$C$ - number of classes, $N_{tot}$- total number of samples,}\\
\multicolumn{7}{l}{$D$ - dimensionality,  Task abr. - task abbreviation in other tables}\\
\multicolumn{7}{l}{$N$ - number of target training samples in task}
\vspace{-15pt}
\end{tabular}}
\end{center}
\end{table} 

\begin{table*}[bt]
\caption{Average Gmean values for different variants of the proposed \ac{REF} algorithm}
\vspace{-5pt}
\label{tab:REFvariants}
\begin{center}
\resizebox{1\linewidth}{!}{
\begin{tabular}{|c||c|c|c|c|c|c|c|c|c|c||c|c|}
\hline
\multicolumn{11}{|c||}{\rule{0pt}{1em} Hyperparameters optimized \rule[-0.5em]{0pt}{1em} } & \multicolumn{2}{|c|}{\rule{0pt}{1em} Default hyperparams \rule[-0.5em]{0pt}{1em} }\\
\hline
& $\abs$ & $\sqr$ & $\cos$-$\abs$ & $\cos$ & $\sin$ & $\tanh$  & $\abs$ & $\cos$-$\abs$ & $\cos$ & base &  & 
\\ \textbf{Data} & $\Lone/D$ & $\Lone/D$ & $\Lone/D$ & $\Lone/D$ & $\Lone/D$ & $\Lone/D$ & $\Ltwo/D$ & $\Ltwo/D$ & $\Ltwo/D$ & $\Lone/D$ & \ac{REF} & base \\
\hline
Iris1 & \bf{93.6$\pm$6.4}& \bf{93.6$\pm$6.4}& \bf{93.6$\pm$6.4}& 76.1$\pm$16.3& 56.4$\pm$8.8& 76.9$\pm$8.5& \bf{93.6$\pm$6.4}& \bf{93.6$\pm$6.4}& 76.7$\pm$15.1& 86.9$\pm$7.5& \bf{93.6$\pm$6.4}& 85.5$\pm$6.2\\ 
Iris2 & \bf{95.6$\pm$1.5}& \bf{95.6$\pm$1.5}& \bf{95.6$\pm$1.5}& 76.3$\pm$10.0& 47.2$\pm$15.7& 44.4$\pm$15.9& \bf{95.6$\pm$1.5}& \bf{95.6$\pm$1.5}& 77.2$\pm$9.9& 87.9$\pm$3.8& \bf{95.6$\pm$1.5}& 84.2$\pm$3.6\\ 
Iris3 & \bf{88.5$\pm$3.8}& \bf{88.5$\pm$3.8}& \bf{88.5$\pm$3.8}& 71.5$\pm$7.7& 48.6$\pm$6.5& 47.8$\pm$8.7& \bf{88.5$\pm$3.8}& \bf{88.5$\pm$3.8}& 73.7$\pm$4.2& 88.0$\pm$4.6& \bf{88.5$\pm$3.8}& 85.5$\pm$10.4\\ 
Seed1 & 82.3$\pm$4.3& 82.3$\pm$4.3& 82.3$\pm$4.3& \bf{85.6$\pm$1.2}& 36.3$\pm$13.0& 28.4$\pm$18.8& 82.3$\pm$4.3& 82.3$\pm$4.3& \bf{85.6$\pm$1.2}& 85.5$\pm$4.2& 82.1$\pm$4.3& 79.2$\pm$4.7\\ 
Seed2 & \bf{90.0$\pm$4.1}& \bf{90.0$\pm$4.1}& \bf{90.0$\pm$4.1}& 83.7$\pm$4.0& 38.9$\pm$8.6& 66.6$\pm$9.1& \bf{90.0$\pm$4.1}& \bf{90.0$\pm$4.1}& 77.7$\pm$15.1& 80.6$\pm$4.0& \bf{90.0$\pm$4.1}& 78.8$\pm$5.0\\ 
Seed3& \bf{93.0$\pm$4.1}& \bf{93.0$\pm$4.1}& \bf{93.0$\pm$4.1}& 86.4$\pm$4.3& 44.6$\pm$16.3& 71.4$\pm$14.5& \bf{93.0$\pm$4.1}& \bf{93.0$\pm$4.1}& 84.1$\pm$7.3& 91.5$\pm$4.0& \bf{93.0$\pm$4.1}& 86.1$\pm$5.8\\ 
Ion1 & \bf{91.6$\pm$2.9}& 90.9$\pm$4.0& 90.9$\pm$4.0& 77.3$\pm$4.6& 54.1$\pm$2.7& 55.3$\pm$7.0& 90.7$\pm$2.9& 91.1$\pm$3.9& 66.2$\pm$4.9& 70.9$\pm$2.5& 91.6$\pm$4.0& 70.6$\pm$2.7\\ 
Ion2 & 57.3$\pm$2.8& 26.9$\pm$8.1& 54.3$\pm$3.9& 38.5$\pm$7.1& 74.5$\pm$6.8& \bf{74.9$\pm$7.3}& 60.8$\pm$3.7& 54.8$\pm$4.3& 7.2$\pm$10.7& 44.9$\pm$4.9& 57.5$\pm$3.2& 17.2$\pm$6.8\\ 
Son1 & 54.9$\pm$2.2& 54.6$\pm$2.1& 54.6$\pm$2.1& 53.6$\pm$2.2& 61.1$\pm$6.3& \bf{65.6$\pm$5.2}& 55.8$\pm$2.4& 54.6$\pm$2.1& 0.0$\pm$0.0& 55.8$\pm$1.5& 55.2$\pm$2.7& 50.7$\pm$3.3\\ 
Son2 & 49.8$\pm$10.0& 51.9$\pm$7.7& 51.9$\pm$7.7& 52.8$\pm$3.9& 47.7$\pm$5.0& 43.2$\pm$3.3& 48.8$\pm$9.9& 51.9$\pm$7.7& 0.0$\pm$0.0& \bf{64.7$\pm$4.3}& 48.3$\pm$9.1& 39.3$\pm$5.8\\ 
Bank1 & 95.7$\pm$2.5& 94.4$\pm$3.5& 94.2$\pm$3.5& 91.9$\pm$5.3& 52.3$\pm$9.1& 52.3$\pm$9.1& 98.2$\pm$2.0& \bf{99.4$\pm$1.4}& 91.6$\pm$5.9& 92.7$\pm$3.2& 91.0$\pm$0.7& 79.7$\pm$3.9\\ 
Bank2 & 90.7$\pm$5.6& 91.3$\pm$5.5& 93.3$\pm$4.5& 85.7$\pm$4.2& 93.3$\pm$3.2& 93.7$\pm$2.8& \bf{94.0$\pm$4.5}& \bf{94.0$\pm$4.5}& 88.9$\pm$3.8& 92.6$\pm$6.0& 93.0$\pm$5.2& 90.3$\pm$5.0\\ 
Happ1 & 36.5$\pm$10.8& 30.7$\pm$9.3& 22.9$\pm$7.9& 28.8$\pm$6.1& 56.5$\pm$4.5& \bf{57.5$\pm$5.3}& 22.9$\pm$7.9& 22.9$\pm$7.9& 29.1$\pm$8.9& 48.3$\pm$2.9& 22.0$\pm$6.5& 39.7$\pm$5.4\\ 
Happ2 & 58.7$\pm$4.4& \bf{61.0$\pm$9.2}& \bf{61.0$\pm$9.2}& 57.8$\pm$9.9& 58.5$\pm$9.1& 58.1$\pm$7.2& \bf{61.0$\pm$9.2}& \bf{61.0$\pm$9.2}& 59.4$\pm$10.5& 58.5$\pm$5.6& 40.8$\pm$6.0& 47.7$\pm$4.5\\ 
\hline
Aver. & \bf{77.0$\pm$4.7}& 74.6$\pm$5.2& 76.2$\pm$4.8& 69.0$\pm$6.2& 55.0$\pm$8.3& 59.7$\pm$8.8& 76.8$\pm$4.8& 76.6$\pm$4.6& 58.4$\pm$7.0& 74.9$\pm$4.2& 74.4$\pm$4.4& 66.8$\pm$5.2\\ 
\hline
\end{tabular}}
\end{center}
\end{table*}

\begin{table*}[tb]
\caption{Average Gmean values for different \acs{OCC} methods}
\vspace{-5pt}
\label{tab:comparisons}
\begin{center}
\resizebox{1\linewidth}{!}{
\begin{tabular}{|c||c|c|c|c|c|c|c|c|c|c|c|}
\hline 
\multicolumn{12}{|c|}{\rule{0pt}{1em} Hyperparameters optimized \rule[-0.5em]{0pt}{1em} }\\
\hline
& \multirow{2}{*}{\ac{REF}} & Linear & Non-linear & \ac{SVDD}& \multirow{2}{*}{\acs{SSVDD}} &Linear  & Non-linear & \multirow{2}{*}{\acs{OCELM}} & \multirow{2}{*}{\acs{OCKELM}} & \multirow{2}{*}{\acs{AAKELM}} & Naive
\\ \textbf{Data} & & \ac{SVDD} & \ac{SVDD} & + \acs{GRM} & & \ac{OCSVM} &  \ac{OCSVM} & & & & Parzen \\
\hline
Iris1 & 93.6$\pm$6.4& 90.6$\pm$8.2& 90.6$\pm$8.2& 90.6$\pm$8.2& 90.0$\pm$6.0& 57.6$\pm$42.4& 94.5$\pm$3.2& \bf{97.3$\pm$2.9}& 96.6$\pm$3.5& 96.6$\pm$3.5& 94.4$\pm$5.5\\ 
Iris2 & \bf{95.6$\pm$1.5}& 90.7$\pm$3.9& 90.1$\pm$3.3& 90.8$\pm$1.6& 90.1$\pm$4.5& 45.6$\pm$15.3& 93.2$\pm$2.4& 87.6$\pm$10.3& 93.5$\pm$1.9& 94.5$\pm$4.0& 93.2$\pm$1.7\\ 
Iris3 & 88.5$\pm$3.8& 88.1$\pm$7.4& 86.7$\pm$6.2& 86.7$\pm$6.7& 86.9$\pm$5.1& 63.5$\pm$11.2& 83.6$\pm$6.9& 64.5$\pm$31.3& \bf{91.1$\pm$2.9}& 88.5$\pm$9.9& 87.3$\pm$4.3\\ 
Seed1 & 82.3$\pm$4.3& 85.7$\pm$3.8& 83.2$\pm$7.0& 84.7$\pm$4.4& 84.9$\pm$3.7& 29.2$\pm$30.1& 84.6$\pm$4.8& 66.4$\pm$7.5& 85.9$\pm$2.7& 86.8$\pm$2.6& \bf{87.9$\pm$3.5}\\ 
Seed2 & 90.0$\pm$4.1& 87.6$\pm$3.9& 86.9$\pm$5.4& 87.1$\pm$5.4& 89.7$\pm$1.4& 19.3$\pm$6.7& 89.7$\pm$6.1& 78.2$\pm$12.7& \bf{91.7$\pm$5.5}& 90.9$\pm$5.1& 90.3$\pm$7.2\\ 
Seed3 & 93.0$\pm$4.1& 90.0$\pm$2.8& 88.5$\pm$3.7& 90.0$\pm$2.8& 89.4$\pm$3.4& 37.0$\pm$40.0& 90.9$\pm$3.9& 60.7$\pm$37.0& 90.0$\pm$2.9& \bf{95.0$\pm$2.6}& 93.3$\pm$2.6\\ 
Ion1 & 91.6$\pm$2.9& 84.0$\pm$2.2& 83.0$\pm$4.2& 84.9$\pm$2.3& 79.7$\pm$4.0& 60.0$\pm$6.0& 89.3$\pm$3.6& 81.4$\pm$4.1& \bf{92.0$\pm$3.3}& 89.8$\pm$3.9& 78.7$\pm$3.3\\ 
Ion2 & 57.3$\pm$2.8& 9.0$\pm$5.4& 10.1$\pm$6.2& 52.7$\pm$16.0& \bf{59.4$\pm$4.9}& 53.5$\pm$10.6& 31.2$\pm$3.0& 50.5$\pm$5.2& 45.8$\pm$6.6& 43.5$\pm$4.4& 33.9$\pm$7.4\\ 
Son1 & 54.9$\pm$2.2& 54.3$\pm$4.3& 54.9$\pm$2.3& 54.3$\pm$3.5& 50.5$\pm$8.5& 56.9$\pm$9.7& 56.5$\pm$4.8& 49.0$\pm$14.7& \bf{61.0$\pm$5.6}& 55.7$\pm$4.0& 57.1$\pm$2.9\\ 
Son2 & 49.8$\pm$10.0& 61.3$\pm$4.3& 62.7$\pm$4.1& 61.6$\pm$4.2& 54.2$\pm$3.8& 52.4$\pm$6.7& 63.4$\pm$4.2& 51.9$\pm$5.5& 61.2$\pm$7.8& \bf{67.2$\pm$4.5}& 48.0$\pm$4.6\\ 
Bank1 & 95.7$\pm$2.5& 97.2$\pm$2.0& 90.9$\pm$7.6& 95.0$\pm$2.8& 94.0$\pm$5.5& 13.6$\pm$30.5& 96.0$\pm$1.1& 91.9$\pm$4.5& 97.8$\pm$3.0& 97.8$\pm$3.0& \bf{98.3$\pm$1.8}\\ 
Bank2 & 90.7$\pm$5.6& 94.3$\pm$3.1& 90.2$\pm$6.9& 92.6$\pm$5.4& 90.9$\pm$7.2& 33.3$\pm$38.9& 94.0$\pm$3.1& 66.3$\pm$42.1& \bf{96.6$\pm$2.0}& \bf{96.6$\pm$2.0}& 94.7$\pm$3.7\\ 
Happ1 & 36.5$\pm$10.8& 43.8$\pm$6.0& 47.2$\pm$11.4& 45.1$\pm$9.2& 44.2$\pm$9.1& \bf{57.2$\pm$9.3}& 41.3$\pm$6.5& 45.5$\pm$5.2& 45.6$\pm$2.6& 44.7$\pm$3.6& 37.0$\pm$7.8\\ 
Happ2 & 58.7$\pm$4.4& \bf{59.6$\pm$9.5}& 55.9$\pm$4.1& 59.3$\pm$9.6& 56.7$\pm$3.9& 43.3$\pm$5.8& 52.1$\pm$4.5& 53.9$\pm$8.8& 57.1$\pm$3.7& 57.5$\pm$8.3& 58.1$\pm$3.6\\ 
\hline
Aver. & 77.0$\pm$4.7& 74.0$\pm$4.8& 72.9$\pm$5.8& 76.8$\pm$5.9& 75.8$\pm$5.1& 44.5$\pm$18.8& 75.7$\pm$4.1& 67.5$\pm$13.7& \bf{79.0$\pm$3.9}& 78.9$\pm$4.4& 75.2$\pm$4.3\\  
\hline 
\multicolumn{12}{|c|}{\rule{0pt}{1em} Fixed default hyperparameters \rule[-0.5em]{0pt}{1em} }\\
\hline
%& \multirow{2}{*}{\ac{REF}} & Linear & Non-linear & \ac{SVDD}& \multirow{2}{*}{\acs{SSVDD}} &  Linear  & Non-linear & \multirow{2}{*}{\acs{OCELM}} & \multirow{2}{*}{\acs{OCKELM}} & \multirow{2}{*}{\acs{AAKELM}} & Naive
%\\ \textbf{Data} & & \ac{SVDD} & \ac{SVDD} & + \acs{GRM} & &\ac{OCSVM} &  \ac{OCSVM} & & & & Parzen \\
%\hline
Iris1 & 93.6$\pm$6.4& 82.9$\pm$9.0& 82.9$\pm$9.0& 82.9$\pm$9.0& 11.0$\pm$10.2& 34.2$\pm$33.3& \bf{94.5$\pm$3.2}& 77.3$\pm$43.3& 93.7$\pm$5.9& 93.7$\pm$5.9& 89.3$\pm$5.9\\ 
Iris2 & \bf{95.6$\pm$1.5}& 83.4$\pm$5.1& 83.4$\pm$5.1& 83.4$\pm$5.1& 84.9$\pm$5.1& 49.0$\pm$8.8& 92.6$\pm$1.5& 69.9$\pm$15.8& 92.5$\pm$2.4& 92.8$\pm$2.0& 93.2$\pm$1.8\\ 
Iris3 & \bf{88.5$\pm$3.8}& 77.0$\pm$13.6& 79.1$\pm$11.0& 79.1$\pm$11.0& 82.7$\pm$5.2& 49.3$\pm$28.2& 86.0$\pm$3.8& 68.6$\pm$38.9& 84.6$\pm$3.7& 84.6$\pm$3.7& 84.7$\pm$4.2\\ 
Seed1& 82.3$\pm$4.3& 85.1$\pm$5.9& 85.2$\pm$4.1& 85.2$\pm$4.1& 84.7$\pm$4.2& 27.2$\pm$26.9& 85.0$\pm$6.1& 54.2$\pm$31.2& 85.9$\pm$3.5& 85.5$\pm$3.3& \bf{86.8$\pm$4.1}\\ 
Seed2 & 90.0$\pm$4.1& 84.1$\pm$2.9& 83.4$\pm$5.6& 84.0$\pm$5.1& 82.5$\pm$2.9& 13.4$\pm$10.5& 87.8$\pm$5.4& 69.8$\pm$39.6& 89.9$\pm$5.2& \bf{90.4$\pm$6.0}& 88.4$\pm$2.9\\ 
Seed3 & \bf{93.0$\pm$4.1}& 84.1$\pm$3.9& 84.4$\pm$2.4& 84.4$\pm$2.4& 62.6$\pm$2.7& 22.7$\pm$28.7& 90.0$\pm$2.8& 73.0$\pm$34.2& 92.1$\pm$2.6& 91.6$\pm$2.9& 90.8$\pm$3.6\\ 
Ion1& \bf{91.8$\pm$3.5}& 84.0$\pm$2.2& 20.6$\pm$4.7& 56.5$\pm$3.5& 0.0$\pm$0.0& 60.4$\pm$5.1& 88.2$\pm$3.0& 53.7$\pm$15.7& 78.8$\pm$3.9& 78.7$\pm$3.7& 78.1$\pm$4.0\\ 
Ion2& \bf{58.2$\pm$2.1}& 3.1$\pm$7.0& 0.0$\pm$0.0& 20.5$\pm$13.2& 42.9$\pm$11.8& 46.6$\pm$8.6& 7.1$\pm$6.8& 14.8$\pm$10.9& 3.1$\pm$7.0& 3.1$\pm$7.0& 24.6$\pm$5.0\\ 
Son1 & 55.2$\pm$2.7& 52.5$\pm$5.3& 0.0$\pm$0.0& 26.3$\pm$14.9& 3.3$\pm$7.4& 54.5$\pm$8.2& \bf{57.7$\pm$5.3}& 32.5$\pm$11.8& 54.7$\pm$3.4& 52.5$\pm$5.7& 55.9$\pm$3.4\\ 
Son2 & 48.3$\pm$9.1& 62.2$\pm$4.2& 0.0$\pm$0.0& 47.4$\pm$4.9& 3.4$\pm$7.7& 52.5$\pm$5.1& \bf{65.1$\pm$4.2}& 25.3$\pm$23.9& 29.3$\pm$5.4& 28.1$\pm$7.8& 30.5$\pm$2.8\\ 
Bank1 & 91.0$\pm$0.7& 94.7$\pm$1.8& 79.5$\pm$21.9& 95.0$\pm$1.8& 77.6$\pm$21.1& 28.1$\pm$38.6& 94.5$\pm$2.7& 68.3$\pm$30.7& 94.2$\pm$3.2& 94.5$\pm$3.0& \bf{95.2$\pm$3.4}\\ 
Bank2 & \bf{93.0$\pm$5.2}& 86.8$\pm$7.4& 91.0$\pm$4.5& 87.2$\pm$6.4& 90.4$\pm$3.1& 12.1$\pm$7.4& 89.7$\pm$4.5& 91.9$\pm$3.2& 90.3$\pm$5.5& 90.3$\pm$5.5& 89.3$\pm$6.5\\ 
Happ1& 22.0$\pm$6.5& 43.5$\pm$5.8& 36.7$\pm$21.8& 43.2$\pm$5.6& 35.5$\pm$21.8& \bf{52.5$\pm$5.5}& 36.8$\pm$9.0& 33.4$\pm$5.1& 30.1$\pm$11.9& 30.1$\pm$11.9& 25.8$\pm$10.1\\ 
Happ2 & 40.8$\pm$6.0& 47.2$\pm$5.3& \bf{50.3$\pm$8.0}& 48.5$\pm$5.2& \bf{50.3$\pm$8.0}& 40.2$\pm$7.8& 38.4$\pm$10.0& 33.8$\pm$23.3& 44.5$\pm$7.9& 44.5$\pm$7.9& 46.3$\pm$9.0\\ 
\hline
Aver. & \bf{74.5$\pm$4.3}& 69.3$\pm$5.7& 55.5$\pm$7.0& 66.0$\pm$6.6& 50.8$\pm$8.0& 38.8$\pm$15.9& 72.4$\pm$4.9& 54.7$\pm$23.4& 68.8$\pm$5.1& 68.6$\pm$5.5& 69.9$\pm$4.8\\ 
\hline
\end{tabular}}
\end{center}
\end{table*}

\subsection{Comparative Methods and Their Hyperparameters}

For comparisons, I used \ac{OCC} methods from different classifier categories. The methods included the common boundary-based approaches \ac{SVDD} \cite{tax2004support} and \ac{OCSVM} \cite{scholkopf1999support} with both linear and non-linear implementations as well as their recent extensions \ac{GESSVDD} \cite{sohrab2023graph} with graph which is equivalent to \ac{SSVDD} \cite{sohrab2018subspace} and \ac{SVDD} with \ac{GRM} \cite{raitoharju2022referencekernel_IJCNN} using random reference vectors (case 2). \ac{OCELM} \cite{leng2015one} was used with both random weights and kernel implementation. From density-based approaches, Naive Parzen density estimation \cite{duin1976choice} was selected as it performed relatively well, e.g., in \cite{irigoien2014towards}. As a reconstruction based approach, I picked \ac{AAKELM} \cite{gautam2017construction},  which uses an autoencoder approach with \acp{ELM}. For all kernel methods, \ac{RBF} kernel, 
$\kappa(\input_i,\input_j) =  \exp  \left(  -\gamma \| \input_i - \input_j\|_2^2  \right)$, was used.

For all the variants of \ac{SVDD} and \ac{OCSVM}, an implementation from LIBSVM library
%\footnotehttps://www.csie.ntu.edu.tw/~cjlin/libsvm/}
was used. For the non-linear \ac{SVDD} variants, the linear \ac{SVDD} implementation was used along with \ac{NPT} \cite{kwak2013NPT}, while the non-linear \ac{OCSVM} relied on the kernel implementation. The implementation of Naive Parzen was obtained from dd-tools.
%\footnote{https://www.tudelft.nl/ewi/over-de-faculteit/afdelingen/intelligent-systems/pattern-recognition-bioinformatics/pattern-recognition-bioinformatics/data-and-software/dd-tools}. 
For \ac{ELM} methods, the implementations provided for \cite{gautam2017construction} were used. All kernel/\ac{NPT} approaches were unified to use the same kernel function implementation to ensure fair comparison with equivalent kernel parameters. 

When hyperparameters were optimized using outlier training data, 
the value for $\gamma$ for the \ac{RBF} kernel was selected from $\gamma = \{10^{-4}, 10^{-3}, 10^{-2}, 10^{-1}, 10^0, 10^1\}$. $C$ in \ac{SVDD}, $\nu$ in \ac{OCSVM}, and \emph{fracej} for Naive Parzen and \ac{ELM} methods were selected from $\{0.05, 0.1, 0.15, 0.2\}$. For \ac{ELM} methods, the regularization coefficient was selected from $reg = \{10^{-4}, 10^{-2}, 10^0, 10^2, 10^4\}$ and for ELM with random weights the number of neurons was selected from $M= \{1,1.5,2\}\times N$, where $N$ is the number of samples. For \ac{SSVDD}, the additional hyperparameters were selected from $\eta = \{10^-5, 10^-4, 10^-3, 10^-2, 10^-1\}$ and $d=\{1, 2, 3, 4, 5, 10, 20\}$. The number of iteration for \ac{SSVDD} was set to 5. For the Naive Parzen classifier, the width parameter $h$ was optimized using the inbuilt maximum likelihood estimation-based approach except for the Ionosphere dataset, where the inbuilt method crashes. There, I simply set $h$ as 0.5 for each element.

When default parameters were used, the following values were applied: $\gamma=10^{-1}$, $C/\nu/fracej=0.1$, $reg=10^{-4}$, $M=1.5\times N$, $\eta=10^{-1}$, $d=3$, $h_i = 0.5$. These values were selected partially based on the default hyperparameters suggested for the algorithms, partially based on our own experiences on most suitable values. They are not guaranteed to be the best overall set of hyperparameters, which is natural given the challenges in hyperparameter optimization.

\subsection{Comparative Results}

The comparative results are provided in Table~\ref{tab:comparisons}. The upper section has results with the hyperparameters optimized, the lower section has results for fixed default hyperparameters. It can be seen that the performance of \ac{REF} is very competitive compared to the computationally much heavier approaches. In terms of the average Gmean over all the tasks and train-test splittings, only \ac{OCKELM} and \ac{AAKELM} outperform \ac{REF}. Some of the compared approaches (linear and non-linear \ac{SVDD}, linear \ac{OCSVM}, and \ac{OCELM}) are even below the base classifier performance given in Table~\ref{tab:REFvariants}.  

The lower part of Table~\ref{tab:REFvariants} shows that the advantages of using \ac{REF} become clearer when the methods are using fixed default parameters. Even though also \ac{REF} performs slightly worse with fixed hyperparameters, the difference is much smaller than for the competing methods.

\section{Conclusions}

This article introduced a simple and easy-to-use algorithm for one-class classification: \acf{REF}. The method has linear time-complexity, but the experiments showed that it performs well compared to other computationally much more demanding methods. The proposed algorithm performs particularly well when used with fixed default parameters. This is an additional benefit that makes \ac{REF} a compelling baseline method for any one-class classification task.

\bibliographystyle{IEEEbib}
\bibliography{bibliography}

\end{document}